\documentclass[letterpaper]{article} % DO NOT CHANGE THIS
\usepackage{aaai2026}  % DO NOT CHANGE THIS
\usepackage{times}  % DO NOT CHANGE THIS
\usepackage{helvet}  % DO NOT CHANGE THIS
\usepackage{courier}  % DO NOT CHANGE THIS
\usepackage[hyphens]{url}  % DO NOT CHANGE THIS
\usepackage{graphicx} % DO NOT CHANGE THIS
\usepackage{natbib}  % DO NOT CHANGE THIS AND DO NOT ADD ANY OPTIONS TO IT
\usepackage{caption} % DO NOT CHANGE THIS AND DO NOT ADD ANY OPTIONS TO IT
\usepackage{algorithm}
\usepackage{algorithmic}

\usepackage{amssymb}
\usepackage{amsmath}
\usepackage{multirow}
\usepackage{graphicx}
\usepackage{booktabs}
\usepackage{arydshln}
\usepackage{array}
\usepackage{subcaption}
\usepackage{tabularx}
\usepackage{arydshln}
\usepackage{newfloat}
\usepackage{listings}
\DeclareCaptionStyle{ruled}{labelfont=normalfont,labelsep=colon,strut=off} % DO NOT CHANGE THIS
\floatstyle{ruled}
\newfloat{listing}{tb}{lst}{}
\floatname{listing}{Listing}
\title{HyCoRA: Hyper-Contrastive Role-Adaptive Learning for Role-Playing}
\title{HyCoRA: Hyper-Contrastive Role-Adaptive Learning for Role-Playing}
\author {
    Shihao Yang\textsuperscript{\rm 1}\equalcontrib,
    Zhicong Lu\textsuperscript{\rm 2}\equalcontrib,
    Yong Yang\textsuperscript{\rm 1}\equalcontrib,
    Bo Lv\textsuperscript{\rm 2},
    Yang Shen\textsuperscript{\rm 1},
    Nayu Liu\textsuperscript{\rm 1}\thanks{Corresponding author.}
}
\affiliations {
    \textsuperscript{\rm 1}Tianjin Laboratory Autonomous Intelligence Technology and Systems,\\ School of
 Computer Science and Technology, Tiangong University\\
    \textsuperscript{\rm 2} University of Chinese Academy of Sciences\\
    \texttt{\{yshihao,nayuliu\}@tiangong.edu.cn}
    
}
\usepackage{bibentry}
\begin{document}

\maketitle

\begin{abstract}
Multi-character role-playing aims to equip models with the capability to simulate diverse roles. Existing methods either use one shared parameterized module across all roles or assign a separate parameterized module to each role. However, the role-shared module may ignore distinct traits of each role, weakening personality learning, while the role-specific module may overlook shared traits across multiple roles, hindering commonality modeling.  In this paper, we propose a novel \textbf{HyCoRA}: \textbf{Hy}per-\textbf{Co}ntrastive \textbf{R}ole-\textbf{A}daptive learning framework, which efficiently improves multi-character role-playing ability by balancing the learning of distinct and shared traits. Specifically, we propose a Hyper-Half Low-Rank Adaptation structure, where one half is a role-specific module generated by a lightweight hyper-network, and the other half is a trainable role-shared module. The role-specific module is devised to represent distinct persona signatures, while the role-shared module serves to capture common traits. Moreover, to better reflect distinct personalities across different roles, we design a hyper-contrastive learning mechanism to help the hyper-network distinguish their unique characteristics. Extensive experimental results on both English and Chinese available benchmarks demonstrate the superiority of our framework. Further GPT-4 evaluations and visual analyses also verify the capability of HyCoRA to capture role characteristics.
\end{abstract}
\begin{figure}[t]
  \includegraphics[width=\columnwidth]{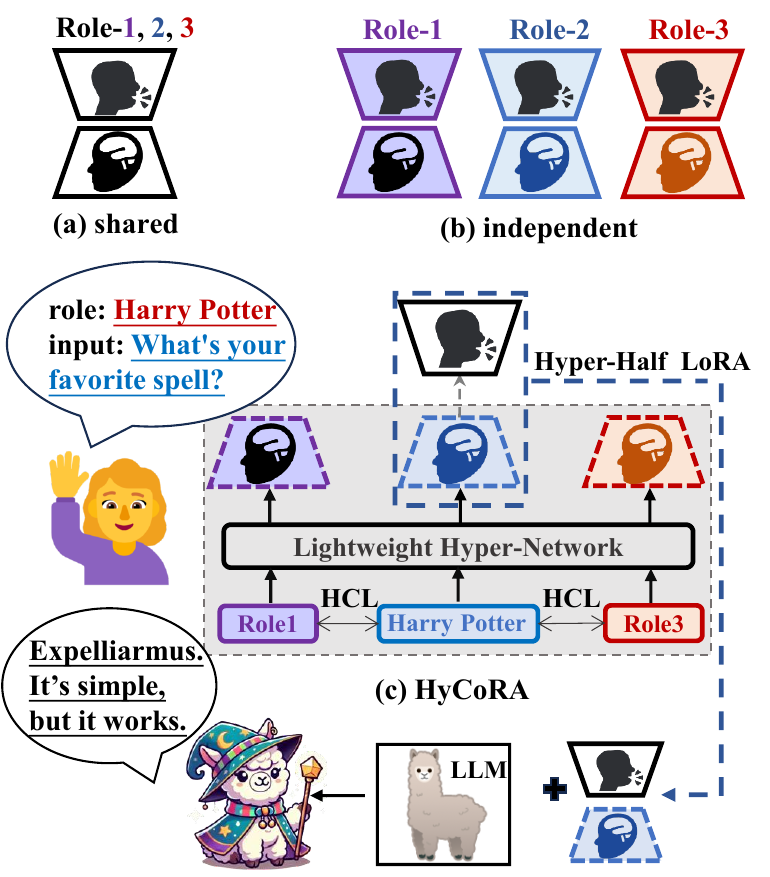}
  \caption{(a) and (b) denote strategies employing a shared module across roles and independent modules assigned to each role, respectively. (c) refers to the HyCoRA proposed in this paper. HCL: Hyper-Contrastive Learning.}
  \label{figure 1}
\end{figure}

% Uncomment the following to link to your code, datasets, an extended version or similar.
% You must keep this block between (not within) the abstract and the main body of the paper.
\begin{links}
    \link{Code}{https://github.com/yshihao-ai/HyCoRA}
    % \link{Datasets}{https://aaai.org/example/datasets}
    % \link{Extended version}{https://aaai.org/example/extended-version}
\end{links}

\section{Introduction}
\label{introduction}
% Role-playing \cite{wu-etal-2025-raiden, tang-etal-2025-rolebreak, castricato-etal-2025-persona, shanahan2023role, tseng-etal-2024-two} allows models to emulate tones, personalities, and role-specific knowledge of multiple characters, improving the user experience by providing immersive and engaging interactions.
The roles in books and movies allow us to witness rich personalities and narratives \cite{lu-etal-2023-narrative}. However,  we are unable to deeply interact with them in real time, which limits opportunities for immersive experiences with roles. Therefore, developing an AI system that can effectively imitate various roles is meaningful. To meet this need, multi-character role-playing (MCRP) \cite{wu-etal-2025-raiden, tang-etal-2025-rolebreak, castricato-etal-2025-persona, tseng-etal-2024-two} has been proposed to deliver an immersive conversational experience in real time by enabling models to simulate diverse characters.

% The instruction-following capability of large language models (LLMs) \cite{achiam2023gpt, touvron2023llama, jiang2023mistral, yang2024qwen2} makes them fit for MCRP. Considering the limited amount of role-playing data, recent approaches often adopt the low-rank adaptation (LoRA) \cite{hu2021lora} method to enhance role-playing ability while preserving the general capabilities of the pre-trained model. Depending on how they capture role characteristics, these methods can generally be grouped into two paradigms: (1) training all characters with a shared adapter \cite{wang-etal-2024-rolellm, tao-etal-2024-rolecraft}, which may fail to adequately capture role-specific traits, as shown in Figure~\ref{figure 1}a; and (2) training each character with an independent adapter \cite{yu-etal-2024-neeko, shao-etal-2023-character}, which not only limits the transfer of common features but also suffers from insufficient training due to the limited data per adapter and incurs extra parameter cost, as shown in Figure~\ref{figure 1}b.

The instruction-following capability of large language models (LLMs) \cite{ grattafiori2024llama, yang2024qwen2, jiang2023mistral,wei-etal-2025-chain} has enabled them to show promising performance in MCRP. Considering the limited amount of role-playing data, recent approaches often adopt the low-rank adaptation (LoRA) \cite{hu2021lora} method to enhance role-playing ability while preserving the general capabilities of the pre-trained model. As shown in Figure~\ref{figure 1}, depending on how they capture role characteristics, existing methods can be broadly grouped into two paradigms: (1)  Using a shared module for all characters \cite{tao-etal-2024-rolecraft, wang-etal-2024-rolellm}; and (2) Assigning an independent module to each character \cite{shao-etal-2023-character, yu-etal-2024-neeko}. Both paradigms have shown clear advantages in capturing role behaviors, but some challenges still remain and warrant further investigation.   Role-shared module may limit the model’s ability to capture fine-grained role-specific traits, due to interference among diverse character features within a single representation space \cite{baziotis-etal-2022-multilingual}. Role-specific module not only limits the transfer of common features across roles but may also suffer from data sparsity for each adapter, leading to insufficient training, while additionally incurring significant parameter overhead.

In this paper, we propose a HyCoRA: Hyper-Contrastive Role-Adaptive Learning framework, which enhances multi-character role-playing ability by balancing the learning of distinct and shared characteristics. Intuitively, different roles interpret questions from distinct views and give answers following shared linguistic rules and common knowledge. To capture both unique and shared behaviors, we introduce a  Hyper-Half LoRA structure, as shown in Figure \ref{figure 1}c. In this structure, we decouple the low-rank adaptation matrices A and B in the LoRA module to serve different functions. Matrix A is generated by a lightweight hyper-network to capture role-specific traits, while Matrix B is a shared trainable matrix to encode common features. Inspired by the hyper-network technique, instead of assigning an independent matrix A to each role, we generate it through a lightweight hyper-network to improve generalization with limited data per role and reduce overall parameters as the number of roles grows. Additionally, to better reflect distinct personalities across different roles, we design a hyper-contrastive learning mechanism to help the hyper-network distinguish their unique traits. It encourages the model to effectively learn each role's traits from its own responses while distinguishing them from traits hidden in other roles' responses.

Extensive experimental results on both English and Chinese available benchmarks demonstrate the superiority of our framework. Furthermore, GPT-4 evaluations and visual analyses indicate HyCoRA's ability to capture distinct characteristics.  In summary, our main contributions include:
\begin{itemize}
\item[$\bullet$]  We propose a  HyCoRA: Hyper-Contrastive Role-Adaptive Learning framework, which efficiently improves multi-character role-playing ability by balancing the learning of distinct and shared characteristics. 
\item[$\bullet$] HyCoRA adopts a Hyper-Half LoRA structure that incorporates a role-specific persona projection module generated by a lightweight hyper-network and a role-shared trait representation module. In addition, we design a hyper-contrastive learning mechanism to better distinguish role-specific characteristics.
\item[$\bullet$] Extensive experiments show that HyCoRA achieves strong performance across both Chinese and English scenarios, as validated by automatic and model-based evaluations. Additionally, we conduct a visualization analysis to provide further insights into the Hyper-Half LoRA structure. The related code will be released to facilitate future research.
\end{itemize}

\begin{figure*}[htbp]
\centering
\includegraphics[width=\textwidth]{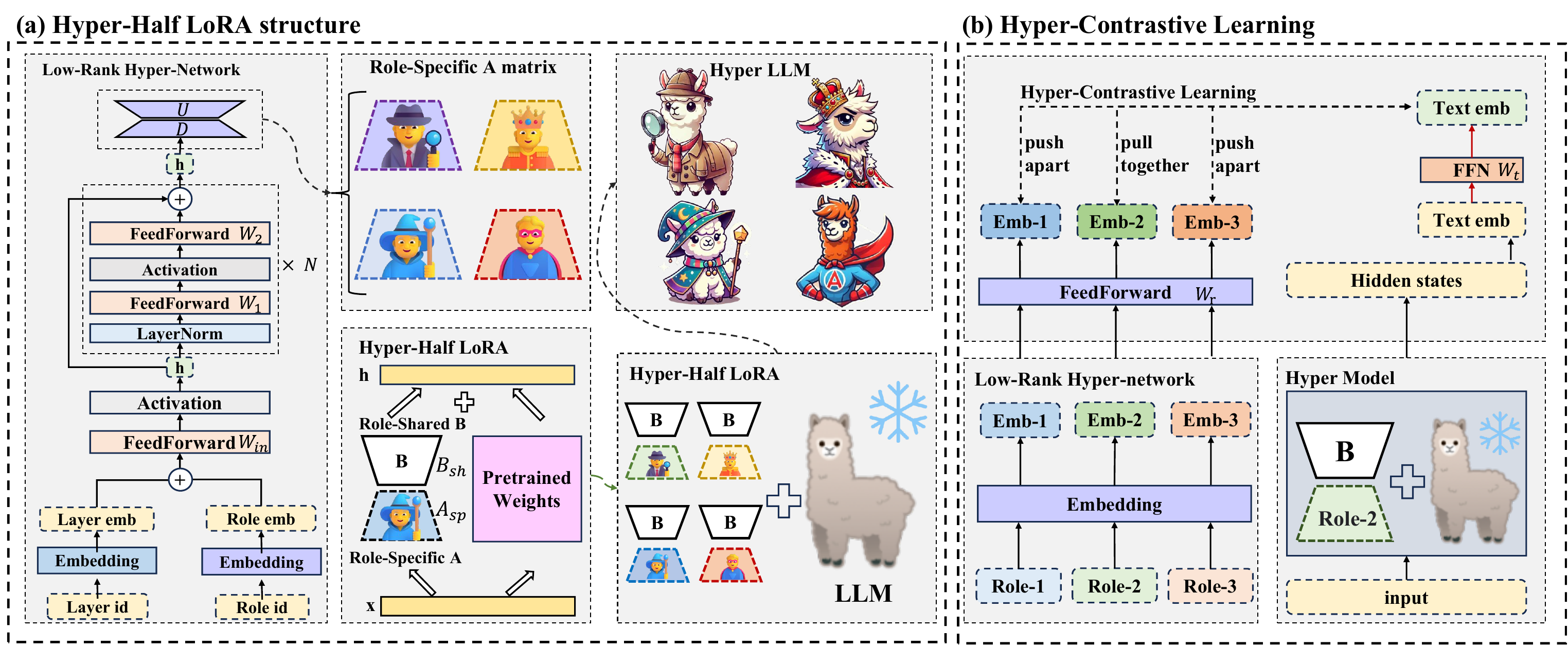}
\caption{An illustration of our framework for MCRP. (a) We construct the Hyper-Half LoRA structure, where the role-specific matrix A is generated by a lightweight hyper-network, and the role-shared matrix B is implemented as a trainable matrix.  (b) We introduce a hyper-contrastive learning mechanism that pulls role representations closer to response representations from the same role and pushes them away from those of different roles.}
\label{figure 2}
\end{figure*}

\section{Related Work}
\subsection{Multi-Character Role-Playing}
Multi-character role-playing (MCRP) \cite{chen-etal-2024-socialbench,tu-etal-2024-charactereval,ran-etal-2024-capturing,chen-etal-2023-large} enables models to emulate diverse roles, allowing them to display various behaviors. Recent research in MCRP has made significant progress \cite{wang-etal-2024-incharacter, wu-etal-2024-role}.  Some methods share parameters across multiple roles, such as DITTO \cite{lu-etal-2024-large}, Character-GLM \cite{zhou-etal-2024-characterglm}, RoleCraft-GLM \cite{tao-etal-2024-rolecraft}, and RoleLLM \cite{wang-etal-2024-rolellm}. These methods can effectively capture common traits shared by different roles, improving generalization.  Other methods allocate distinct parameters to each role, such as Character-LLM \cite{shao-etal-2023-character} and Neeko \cite{yu-etal-2024-neeko}, which helps maintain each role’s unique personality. To balance model size and learning ability, both types of methods often use Low-Rank Adaptation (LoRA)  techniques \cite{hu2021lora} to effectively simulate diverse roles. 
% A detailed introduction to LoRA is provided in Appendix C. 
Despite the improvements made by these methods, there remain several challenges to overcome.  Methods that share parameters may overlook the unique traits of each role, which may weaken personality learning. In contrast, methods with distinct parameters may ignore common traits across roles, limiting the modeling of shared knowledge. In this work, we address these challenges by balancing shared and distinct traits learning to enhance MCRP capabilities.

\subsection{Hyper-Network}
Hyper-networks \cite{liu-etal-2025-language,lv-etal-2024-hyperlora,li-etal-2024-hypernetwork, baziotis-etal-2022-multilingual} have been widely adopted in multi-task \cite{ivison-etal-2023-hint, 10643687, liuenvironment,wei-etal-2023-guide,liu-etal-2025-sara} learning scenarios. Motivated by their ability to dynamically generate parameters for target networks, we view MCRP as a multi-task learning problem, with each role treated as a distinct task. In HyCoRA, rather than training independent parameters for each role with limited role-specific data, the hyper-network is trained across data from all roles, which not only reduces parameter overhead but also addresses insufficient training caused by limited data for each role. Moreover, traditional hyper-networks incur prohibitive computational costs when applied to LLMs. To address this issue, we propose a lightweight hyper-network that notably reduces parameter cost. Additionally, we employ hyper-contrastive learning to help the lightweight hyper-network distinguish features of different roles.

\section{Methodology}
\label{Methodology}
In this section, we detail our proposed  HyCoRA framework for MCRP, which balances the learning of distinct and shared traits.  As shown in Figure \ref{figure 2}a,  HyCoRA adopts a Hyper-Half LoRA structure, which consists of a role-specific persona projection module generated by a lightweight hyper-network and a role-shared trait representation module implemented as a trainable matrix. The role-specific module aims to represent distinct persona signatures, while the role-shared module serves to capture common features. Incorporating this structure into the backbone LLM yields the Hyper Model. During training, as shown in Figure \ref{figure 2}b, we employ hyper-contrastive learning as an auxiliary task, which utilizes role-specific responses to further help the lightweight hyper-network capture and distinguish distinct characteristics. 
\subsection{Hyper-Half LoRA}
% The Hyper-Half LoRA structure assigns the low-rank matrix A to the role-specific persona projection module and the low-rank matrix B to the role-shared linguistic representation module.

\noindent \textbf{Role-Specific A}\indent The lightweight hyper-network leverages character IDs to generate role-specific matrices A. Additionally, since the matrices should differ across layers even for the same character, layer IDs are introduced to generate layer-specific matrices, as shown in Figure \ref{figure 2}a. We use $c$ and $l$ to denote the character ID and layer ID, respectively. The character and layer embeddings, $e_c \in \mathbb{R}^{d_c}$ and $e_l \in \mathbb{R}^{d_l}$, are obtained by embedding $c$ and $l$ using separate embedding matrices. Then, we concatenate both embeddings and project them with $W_{in}\in \mathbb{R}^{d_h \times (d_c + d_l)}$ to obtain the hyper-network hidden state $h_0 \in \mathbb{R}^{d_h}$. Finally, we pass $h_0$ through $N$ residual blocks to get $h_i\in \mathbb{R}^{d_h}$, which aims to capture high-level features:
% \begin{small}
\begin{align}
&h_0 = ReLU(W_{in}([e_c;e_l])+b_{in}), \\
&h_{i + 1} = W_2(ReLU(W_1h_{i}+b_1))+b_2,
\end{align}
% \end{small}
where $W_{1}\in \mathbb{R}^{d_h\times d_h}$ and $W_{2}\in \mathbb{R}^{d_h\times d_h}$ are the trainable weights of each residual block. 

In traditional hyper-networks, the final hidden state $h_N$ is projected by $W_{A} \in \mathbb{R}^{(r \times d) \times d_h}$, where $(r, d)$ corresponds to the shape of the role-specific matrix $A_{sp}$. However, as $d_h$ increases, the parameter size of $W_A$ grows significantly. To address this challenge, we decompose $W_A$ into two low-rank matrices: $D \in \mathbb{R}^{k \times d_h}$ and $U \in \mathbb{R}^{(r \times d) \times k}$, where $k \ll \min(d_h,\ r \times d)$. The role-specific matrix $A_{sp}$ is computed as follows:
\begin{align}
 A_{sp} = U(D h_N + b_d) + b_u.
\end{align}
As $d_h$ increases, the size of $U$ remains unchanged,  effectively limiting the overall parameter growth. 
% The comparison of parameters between the lightweight and traditional hyper-networks is provided in Appendix F.

The generation process of role-specific matrices A illustrates that the lightweight hyper-network is trained on data from all characters, which ensures sufficient training of the hyper-network. Additionally, in order to generate informative and discriminative role-specific matrices A, the role embedding $e_c$ is required to model the unique traits and distinguish between different characters. Therefore, we will use the hyper-contrastive learning mechanism at a later stage to help the lightweight hyper-network better differentiate between the roles.

\noindent \textbf{Role-Shared B}\indent In contrast to the Role-Specific matrix $A_{sp} \in \mathbb{R}^{r \times d}$, the Role-Shared matrix $B_{sh}  \in \mathbb{R}^{m \times r}$  is a trainable matrix, rather than being generated by the lightweight hyper-network.  Similar to LoRA, we initialize separate trainable matrices $B_{sh}$ for different layers. However, within the same layer, multiple roles share a single matrix $B_{sh}$. Let $W_0 \in \mathbb{R}^{m \times d}$ represent the weight matrix of a pretrained LLM. For the original $out={W}_\mathrm{0}x$, the modified forward pass in the Hyper-Half LoRA structure is given by:
\begin{align}
{out}={W}_{0}x+\frac{\alpha}{r}{B}_{sh}{A}_{sp}x,
\end{align}
where $\alpha$ is a scaling factor, and $r$  is the decomposition rank. 

Multiple roles share the matrix \( B_{sh} \), which not only reduces the number of trainable parameters compared to fully role-specific decompositions, but also helps the model capture common features across roles. When integrated into the Hyper-Half LoRA structure, this role-shared component allows the lightweight hyper-network to focus on modeling role-specific traits via \( A_{sp} \).

\subsection{Hyper-Contrastive Learning}
A role’s preferences and personality are often implicitly reflected in its responses. Since each role is represented by an embedding in the lightweight hyper-network, these responses can serve as useful signals to help the hyper-network learn more discriminative role embeddings.  Motivated by this, we propose a hyper-contrastive learning mechanism, which leverages character responses to help the lightweight hyper-network better distinguish role-specific traits.
  
We refer to an LLM equipped with the Hyper-Half LoRA structure as a Hyper Model. As illustrated in Figure~\ref{figure 2}b, for each batch, role IDs are embedded via the lightweight hyper-network to produce role embeddings. Simultaneously, the Hyper Model processes the question-answer pairs to generate token-level hidden states. To capture the overall semantics of the response, we extract the hidden state of the final token as the sequence representation. This final state and the role embeddings are separately projected into an $m$-dimensional space, yielding the role and response representations:
\begin{align}
&z_{r_i} = W_{r}e_{c_i}+b_{r}, \ \ z_{t_i} = W_{t}e_{t_i}+b_{t},
\end{align}
where $e_{c_i}$ is the role embedding of the role in the $i$-th sample, $e_{t_i}$ is the final token’s hidden state of the corresponding response, and $W_r, W_t, b_r, b_t$ are trainable projection parameters.

To encourage the lightweight hyper-network better distinguish role-specific characteristics, we define a hyper-contrastive learning loss.  For the $i$-th sample, the role representation is encouraged to be close to the response representations in the batch that have the same role as $i$-th sample  (i.e., positive samples), and to be distant from those of samples with different roles (i.e., negative samples).  The cross-entropy loss and the hyper-contrastive loss together constitute the HyCoRA's loss function. However, due to the randomized state of the hyper-network parameters in the early stages of training, the final token hidden state cannot effectively represent the text sequence. Therefore, only the cross-entropy loss is initially used. Once the model begins to capture basic sentence semantics, the hyper-contrastive loss is introduced. The overall loss function is defined as:

\begin{small}
\begin{align}
&\mathcal{L} = - \sum_{i=1}^{n} \log P(y_i \mid x_{\leq i}) + \lambda \mathcal{L}_{{cl}}, \\
&\mathcal{L}_{{cl}} = \sum_{i \in {I}} \frac{-1}{|P(i)|} \sum_{p \in P(i)} \log \frac{e^{\mathrm{sim}(z_{r_i}, z_{t_p}) / \tau}}{\sum_{j \in I} e^{\mathrm{sim}(z_{r_i}, z_{t_j}) / \tau}}, \\
&\lambda = \lambda_\mathrm{min} + (\lambda_\mathrm{max} - \lambda_\mathrm{min}) \cdot \max\left(0, \frac{ep - ep_\mathrm{st}}{ep_\mathrm{to} - ep_\mathrm{st}}\right),
\end{align}
\end{small}
where $\mathrm{sim}(p, q)$ denotes cosine similarity, $I$ is the set of samples in a batch, and $P(i)$ represents the subset of samples in $I$ that have the same role as the $i$-th sample. $z_{r_i}$ and $z_{t_p}$ are the role and response representations, respectively. $\lambda$ controls the weight of the hyper-contrastive loss, increasing linearly from $\lambda_\mathrm{min}$ to $\lambda_\mathrm{max}$ between epoch $ep_\mathrm{st}$ and $ep_\mathrm{to}$.
\begingroup
\setlength{\tabcolsep}{0.8mm}
\begin{table*}[!ht]
  {\small
  \centering
      \begin{tabular}{lcccccccc}
        \toprule[1pt]
        \multirow{2}{*}{\textbf{Models}} & \multicolumn{4}{c}{\textbf{General}} & \multicolumn{4}{c}{\textbf{Specific}} \\
        \cmidrule(lr){2-5} \cmidrule(lr){6-9}
        & \textbf{BLEU $\uparrow$} & \textbf{ROUGE-1 $\uparrow$} & \textbf{ROUGE-2 $\uparrow$} & \textbf{ROUGE-L $\uparrow$} & \textbf{BLEU $\uparrow$} & \textbf{ROUGE-1 $\uparrow$} & \textbf{ROUGE-2 $\uparrow$} & \textbf{ROUGE-L $\uparrow$} \\
        \midrule
         GPT-4 & 37.02 & 58.55 & 36.53 & 52.14 & 07.32 & 35.57 & 12.28 & 23.06 \\
        \midrule
        ChatPLUG & 16.14 & 40.43 & 19.50 & 34.63 & 08.14 & 36.33 & 13.44 & 24.53 \\
        Character.AI & 15.76 & 44.92 & 20.71 & 36.45 & 08.74 & 37.17 & 13.91 & 25.58 \\
        Yi-6B-Chat & 16.55 & 40.21 & 19.82 & 34.17 & 06.74 & 33.03 & 11.68 & 21.99 \\
        \midrule
        ChatGLM2 & 14.83 & 39.48 & 18.73 & 32.91 & 07.74 & 36.08 & 12.82 & 23.15 \\
        ChatGLM2-LoRA (RoleLLM) & 34.45 & 56.48 & 34.29 & 50.22 & 12.09 & 42.62 & 18.66 & 30.14 \\
        ChatGLM2-HyCoRA & \textbf{35.09} & \textbf{56.65} & \textbf{34.56} & \textbf{50.69} & \textbf{13.40} & \textbf{43.11} & \textbf{19.78} & \textbf{30.71} \\
        \midrule
        Qwen2-Instruct & 17.25 & 43.98 & 20.30 & 36.14 & 05.20 & 32.13 & 08.95 & 20.21 \\
        Qwen2-LoRA (RoleLLM) & 35.54 & 56.85 & 34.84 & 50.60 & 13.86 & 44.19 & \textbf{20.63} & 31.92 \\
        Qwen2-HyCoRA & \textbf{35.80} & \textbf{57.16} & \textbf{35.38} & \textbf{51.18} & \textbf{13.88} & \textbf{44.34} & 20.50 & \textbf{32.02}\\
        \midrule
        Qwen2.5-Instruct & 19.07 & 46.07 & 22.27 & 37.68 & 04.63 & 31.25 & 08.37 & 19.43\\
        Qwen2.5-LoRA (RoleLLM) & 34.18 & 56.21 & 34.01 & 50.25 & 14.15 & 44.07 & 20.82 & 31.65 \\
        Qwen2.5-HyCoRA & \textbf{34.80} & \textbf{56.78} & \textbf{34.49} & \textbf{50.65} & \textbf{14.21} & \textbf{44.23} & \textbf{20.92} & \textbf{32.51} \\
        \bottomrule[1pt]
      \end{tabular}
    }
    \caption{Model performance assessment in the Chinese scenario based on automatic evaluation.}
  \label{table 1}
\end{table*}
\endgroup

\begingroup
\setlength{\tabcolsep}{0.9mm}
\begin{table*}[!ht]
  {\small
  \centering
      \begin{tabular}{lcccccccc}
        \toprule[1pt]
        \multirow{2}{*}{\textbf{Models}} & \multicolumn{4}{c}{\textbf{General}} & \multicolumn{4}{c}{\textbf{Specific}} \\
        \cmidrule(lr){2-5} \cmidrule(lr){6-9}
        & \textbf{BLEU $\uparrow$} & \textbf{ROUGE-1 $\uparrow$} & \textbf{ROUGE-2 $\uparrow$} & \textbf{ROUGE-L $\uparrow$} & \textbf{BLEU $\uparrow$} & \textbf{ROUGE-1 $\uparrow$} & \textbf{ROUGE-2 $\uparrow$} & \textbf{ROUGE-L $\uparrow$} \\
        \midrule
        GPT-4 & 42.67 & 55.03 & 32.62 & 53.56 & 08.01 & 32.04 & 09.89 & 28.83 \\
        \midrule
        ChatPLUG & 10.00 & 24.01 & 10.57 & 22.57 & 04.24 & 25.45 & 05.48 & 22.67 \\
        Character.AI & 17.66 & 30.03 & 14.46 & 28.18 & 05.05 & 25.74 & 06.45 & 23.27 \\
        Vicuna-13B & 18.67 & 34.88 & 15.80 & 33.50 & 06.30 & 29.05 & 08.14 & 26.06 \\
        Alpaca-7B & 13.52 & 27.58 & 13.66 & 26.79 & 07.06 & 31.17 & 09.87 & 27.89 \\
        \midrule
        LLaMA-2-7B-Chat & 11.72 & 28.60 & 12.21 & 26.88 & 02.33 & 20.55 & 04.29 & 18.81 \\
        LLaMA-2-LoRA (RoleLLM) & 25.50 & 38.68 & 19.97 & 37.11 & 09.95 & 34.63 & 12.02 & 31.14 \\
        LLaMA-2-HyCoRA & \textbf{26.26} & \textbf{39.29} & \textbf{21.23} & \textbf{37.65} & \textbf{10.12} & \textbf{34.85} & \textbf{12.15} & \textbf{31.31} \\
        \bottomrule[1pt]
        \end{tabular}
    }
    \caption{Model performance assessment in the English scenario based on automatic evaluation.}
  \label{table 2}
\end{table*}
\endgroup
\section{Experiments}
% In this section, we first present the datasets, the evaluation metrices used in the experiments , and the baselines. Then, we present the experimental results and offer a detail analysis.

\subsection{Datasets}
To verify the effectiveness of our framework in MCRP, we use RoleBench \cite{wang-etal-2024-rolellm}  as the benchmark dataset. RoleBench contains 168,093 samples across 100 distinct roles (5 in Chinese and 95 in English), covering both general and role-specific knowledge. General knowledge assesses the model’s ability to capture tone and personality, while role-specific knowledge evaluates its understanding of the character's background.

\subsection{Implementation Details}
For comprehensive evaluation across backbone models, we adopt several popular LLMs. For the Chinese scenario, we use ChatGLM2-6B\footnote{https://huggingface.co/THUDM/chatglm2-6b} \cite{zeng2022glm}, Qwen2-7B\footnote{https://huggingface.co/Qwen/Qwen2-7B} \cite{bai2023qwen}, and Qwen2.5-7B\footnote{https://huggingface.co/Qwen/Qwen2.5-7B}\cite{yang2024qwen2} as backbone models. For the English scenario, we adopt LLaMA-2-7B\footnote{https://huggingface.co/meta-llama/Llama-2-7b-hf} \cite{touvron2023llama} as the backbone model. Specifically, LLaMA-2-7B, Qwen2-7B, and Qwen2.5-7B are pre-trained models, while ChatGLM2-6B  is post-trained with enhanced dialogue capabilities. 
% Detailed hyperparameter configurations for HyCoRA are provided in the Appendix D.

\subsection{Evaluation metrics}
To evaluate our framework from multiple perspectives, we employ both automatic and model-based evaluation methods. Following previous research \cite{chen2024oscars, tao-etal-2024-rolecraft, wang-etal-2024-rolellm, yu-etal-2024-neeko}, we use ROUGE \cite{lin2004rouge} and BLEU \cite{papineni2002bleu} as automatic metrics to quantitatively assess output quality. For model-based evaluation, we use a ranking-based approach to calculate win rates. Specifically, GPT-4 assesses the model's performance across two dimensions: \textbf{(1) Role Behavior (RB)}, which examines whether the responses align with the character's tone and personality traits, and \textbf{(2) Role Knowledge (RK)}, which evaluates the consistency of responses with the character's background knowledge. 
% The detailed GPT-4 evaluation prompt is provided in Appendix J.
\subsection{Baselines}
(1) RoleLLM \cite{wang-etal-2024-rolellm} adopts a shared LoRA module across roles, and the same strategy is also used in
RoleCraft-GLM \cite{tao-etal-2024-rolecraft}. (2) Independent LoRA (Ind. LoRA) assigns an independent LoRA module to each role. 
(3) Neeko \cite{yu-etal-2024-neeko} adopts a similar approach to Ind.LoRA but introduces an additional gating mechanism. 
Due to cost constraints, (2) and (3) are only compared in the Chinese scenario. (4) Role-playing models are used as baselines due to their strong role-playing capabilities, including ChatPLUG \cite{tian2023chatplug} and Character.AI\footnote{https://character.ai/}. (5) Instruction-tuned models are used as baselines due to their excellent instruction-following capabilities. These models include LLaMA-2-7B-Chat \cite{touvron2023llama}, Vicuna-13B \cite{chiang2023vicuna}, and Alpaca-7B \cite{alpaca} for the English scenario, as well as ChatGLM2-6B, Yi-6B-Chat\footnote{https://huggingface.co/01-ai/Yi-6B-Chat}, Qwen2-7B-Instruct \cite{bai2023qwen}, and Qwen2.5-7B-Instruct \cite{yang2024qwen2} for the Chinese scenario. (6) GPT-4 is also selected to serve as an upper-bound reference model. 

\subsection{Performance Analysis}
% \subsubsection{automatic evaluation}
\noindent \textbf{Automatic Evaluation Results}
are reported in Tables \ref{table 1} and \ref{table 2}. Overall, HyCoRA outperforms other baselines, reflecting its superior ability to simulate diverse roles. Compared with role-playing and instruction-tuned models, LoRA and HyCoRA demonstrate notable advantages, which reveal the challenges of capturing diverse character traits in multi-character role-playing. Compared with LoRA, HyCoRA shows more marked advantages in the Chinese scenario than in the English scenario. We attribute this to HyCoRA generating role-specific matrices directly using role IDs, without requiring role-specific personality instructions, which lowers token consumption. This indicates that HyCoRA learns character traits from role responses. However, as the number of roles increases, similar responses across roles become inevitable, which may confuse the model. Therefore, we speculate that providing HyCoRA with role-specific personality instructions, similar to those used in LoRA, could potentially enhance its performance. Furthermore, HyCoRA outperforms GPT-4 in some scenarios but falls short in others, which may be attributed to GPT-4's large parameter size.
\begingroup
\setlength{\tabcolsep}{1mm}
\begin{table}[!ht]
  {\small
  \centering
    \begin{tabular}{l c c c c c c c}
        \toprule[1pt]
        \textbf{Methods} & \textbf{Datasets} & \textbf{Role-1} & \textbf{Role-2} & \textbf{Role-3} & \textbf{Role-4} & \textbf{Role-5} \\
        \midrule
        Ind. LoRA & General & 50.30 & 42.85 & 49.46 & 49.81 & 48.38  \\
        Neeko & General & 51.02 & 42.31 & 52.35 & 50.32 & 48.32 \\
        HyCoRA    & General & \textbf{53.58} & \textbf{43.73} & \textbf{54.33} & \textbf{52.66} & \textbf{50.13} \\
        \midrule
        Ind. LoRA & Specific & 25.46 & 27.04 & 30.40 & 27.77 & \textbf{31.97} \\
        Neeko & Specific & 30.41 & 28.22 & 31.20 & 27.72 & 31.12 \\
        HyCoRA           & Specific & \textbf{32.56} & \textbf{31.53} & \textbf{34.19} & \textbf{29.51} & 30.88 \\
        \bottomrule[1pt]
      \end{tabular}
    }
  \caption{ROUGE-L evaluation results for Independent LoRA, Neeko, and HyCoRA.}
  \label{table:Independent LoRA}
\end{table}
\endgroup

\noindent \textbf{Comparison with Ind. LoRA and Neeko}. To further assess model performance, we conduct experiments to evaluate the imitation ability of each role, as shown in Table~\ref{table:Independent LoRA}. Specifically, we report ROUGE-L scores for each role in the Chinese scenario using Qwen2-7B as the backbone model. The results show that HyCoRA outperforms both baselines in most cases. Since both baselines adopt a strategy of assigning independent LoRA modules to each role, we hypothesize that the limited amount of role-specific training data prevents these modules from being sufficiently trained, which may impair their generalization during evaluation.

\begingroup
\setlength{\tabcolsep}{3.0mm}
\begin{table}[!ht]
  \centering
  {\small
    \begin{tabular}{l l c c c}
        \toprule[1pt]
        \textbf{Models} & \textbf{Methods} & \textbf{RB $\uparrow$} & \textbf{RK $\uparrow$} & \textbf{avg. $\uparrow$} \\
        \midrule
        \multirow{2}{*}{ChatGLM2-6B} & LoRA & 0.33 & 0.38 & 0.36 \\
                                  & HyCoRA & \textbf{0.52} & \textbf{0.59} & \textbf{0.56} \\
        \midrule
        \multirow{2}{*}{Qwen2-7B} & LoRA & 0.36 & 0.39 & 0.38 \\
                                & HyCoRA & \textbf{0.46} & \textbf{0.57} & \textbf{0.52} \\
        \midrule
        \multirow{2}{*}{Qwen2.5-7B} & LoRA & 0.41 & 0.42 & 0.42 \\
                                & HyCoRA & \textbf{0.48} & \textbf{0.54} & \textbf{0.50} \\
        \midrule
        \multirow{2}{*}{LLaMA-2-7B} & LoRA & 0.25 & \textbf{0.38} & 0.32 \\
                                & HyCoRA & \textbf{0.35} & 0.37 & \textbf{0.36} \\
        \bottomrule[1pt]
      \end{tabular}
  }
  \caption{Results of the GPT-4-based evaluation for LoRA and HyCoRA on backbone models.}
  \label{table:gpt4_evaluation}
\end{table}
\endgroup
% \subsubsection{GPT-4-Based Evaluation}

\noindent \textbf{GPT-4-Based Evaluation Results} are reported in Table \ref{table:gpt4_evaluation}. HyCoRA demonstrates competitive performance in both the Role Behavior (RB) and Role Knowledge (RK) metrics. Specifically, the advantage in RB suggests that HyCoRA tends to better capture each role’s tone and personality in its responses. In RK, it shows a stronger grasp of role-specific knowledge, effectively reflecting role-relevant information. These results suggest that HyCoRA can maintain character authenticity while conveying accurate knowledge.

\begin{figure}[!ht]
  \includegraphics[width=\columnwidth]{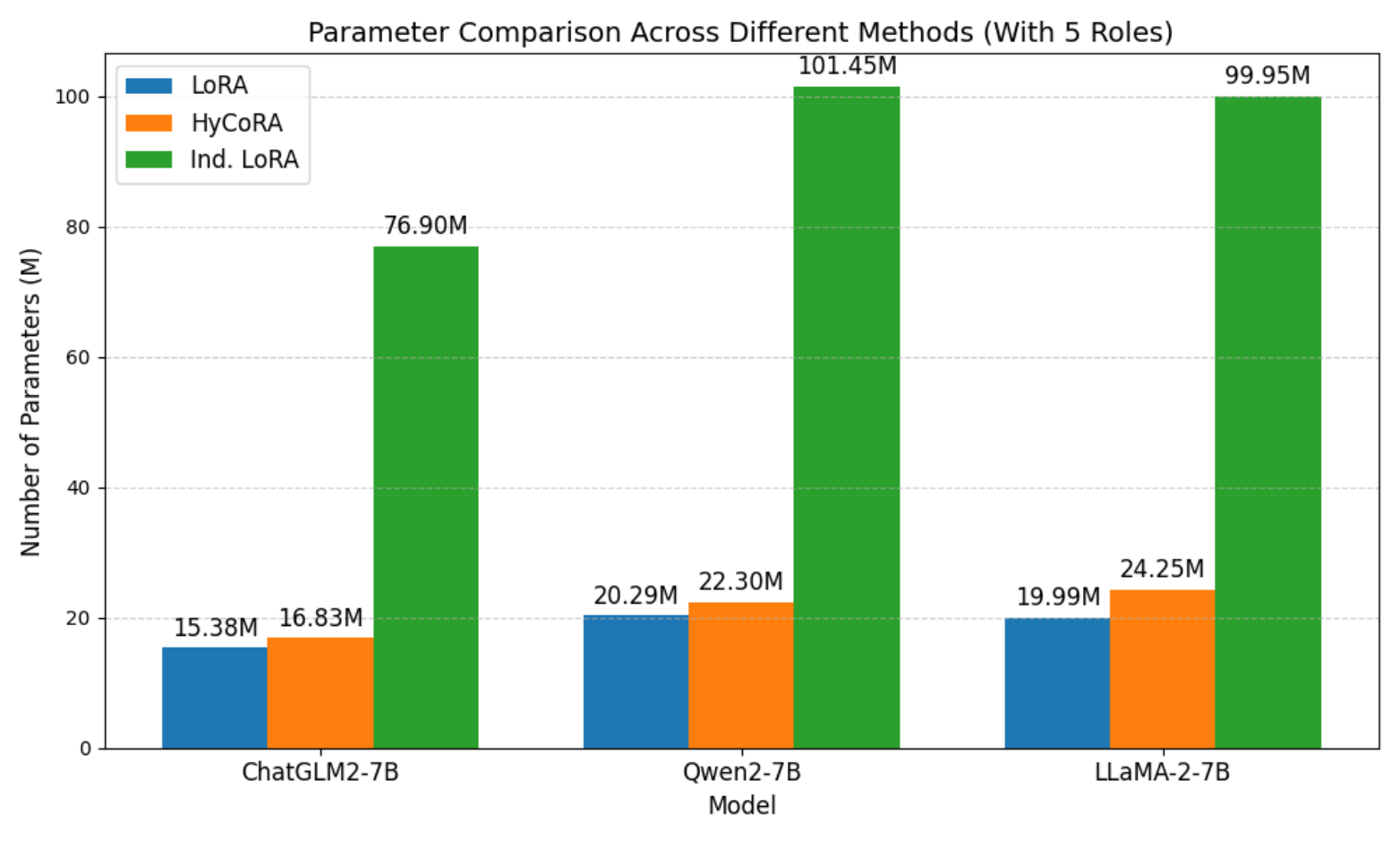}
  \caption{The trainable parameters for different methods.}
  \label{parameters of LoRA}
\end{figure}
\noindent \textbf{Trainable Parameter Analysis.} Finally, we compared the number of trainable parameters across different methods. As shown in Figure \ref{parameters of LoRA}, HyCoRA reduces the trainable parameters compared to the independent LoRA method, while being closer to the shared LoRA method.

\subsection{Ablation Analysis}
To comprehensively assess the effectiveness of our framework, we set up the following ablation variants. \textbf{(1) HyCoRA (w/o HCL)} removes the hyper-contrastive learning mechanism. \textbf{(2) Rsp A \& Rsp B} uses role-specific matrices for both A and B. \textbf{(3) Mrs A \& Rsp B} adopts a shared matrix for A and a role-specific matrix for B. \textbf{(4) Rsp A \& Mrs B (i.e., HyCoRA)}  applies a role-specific matrix for A and a shared matrix for B. In these variants, the shared matrices are initialized as trainable parameters, while the role-specific matrices are generated by the lightweight hyper-network.

\begingroup
\setlength{\tabcolsep}{1.0mm}
\begin{table}[!ht]
  \centering
  {\small
      \begin{tabular}{lccc}
        \toprule[1pt]
        \textbf{Models} & \textbf{Datasets} & \textbf{BLEU-4} & \textbf{ROUGE-L} \\
        \midrule
        HyCoRA (w/o HCL) & General & 34.64 & 50.42 \\
        HyCoRA & General & \textbf{35.09} & \textbf{50.69} \\
        \midrule 
        HyCoRA (w/o HCL) & Specific  & 12.65 & 30.34 \\
        HyCoRA & Specific  & \textbf{13.40} & \textbf{30.71} \\
        \midrule
        Rsp A \& Rsp B & General & 32.60 & 48.89  \\
        Mrs A \& Rsp B &  General & 33.81 & 49.78 \\
        Rsp A \& Mrs B (HyCoRA) &  General & \textbf{35.09} & \textbf{50.69} \\
        \midrule
        Rsp A \& Rsp B &  Specific & 12.96 & 30.02  \\
        Mrs A \& Rsp B & Specific & 12.74 & 30.25 \\
        Rsp A \& Mrs B (HyCoRA) & Specific & \textbf{13.40} & \textbf{30.71}\\
        \bottomrule[1pt]
      \end{tabular}
    }
    \caption{Comparative results of HyCoRA with and without HCL, and different combinations of role-specific (\textbf{Rsp}) and multi-role-shared (\textbf{Mrs}) matrices.}
    \label{table 4}
\end{table}
\endgroup

To validate the effectiveness of the hyper-contrastive learning (HCL) mechanism, we conduct comparative experiments using ChatGLM2-6B as the backbone model. As shown in Table \ref{table 4}, removing the HCL mechanism leads to a decline in model performance across both scenarios, indicating that HCL enhances the model's ability to capture role-specific traits. We also study different Hyper-Half LoRA configurations. Table \ref{table 4} shows that using two role-specific matrices performs worst, emphasizing the need for a shared matrix. This allows the role-specific matrix to focus on capturing unique traits. The Hyper-Half LoRA setup (Rsp A \& Mrs B) achieves the best results, effectively modeling both shared and role-specific features.

\begin{figure*}[ht]
    \centering
    \begin{subfigure}[b]{0.32\textwidth}
        \includegraphics[width=\textwidth]{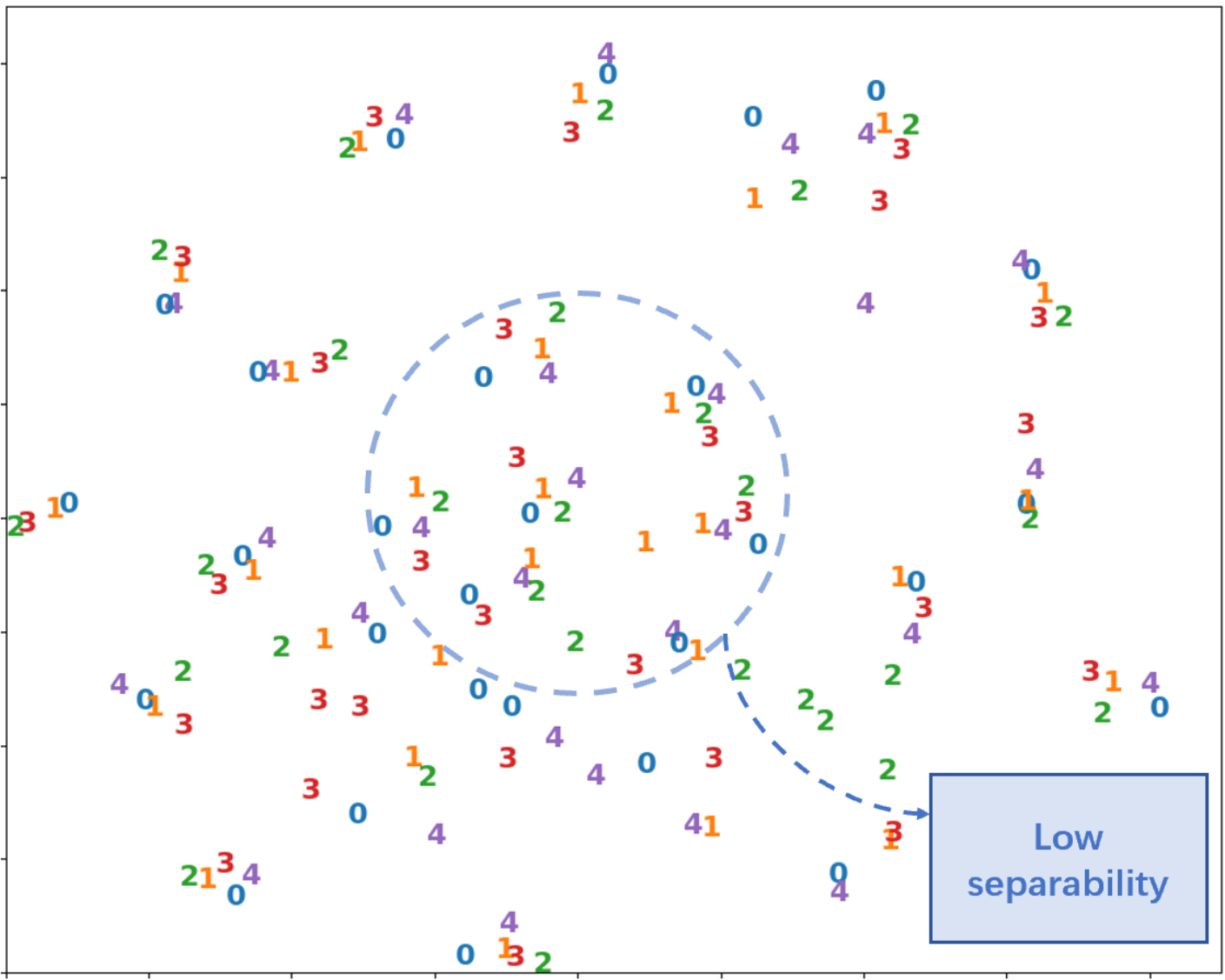}
        \caption{\textbf{Rsp A \& Rsp B: } The distribution in the vector space when combining the role-specific matrix A and the role-specific matrix B.}
        \label{fig:tsne1}
    \end{subfigure}
    \hfill
    \begin{subfigure}[b]{0.32\textwidth}
        \includegraphics[width=\textwidth]{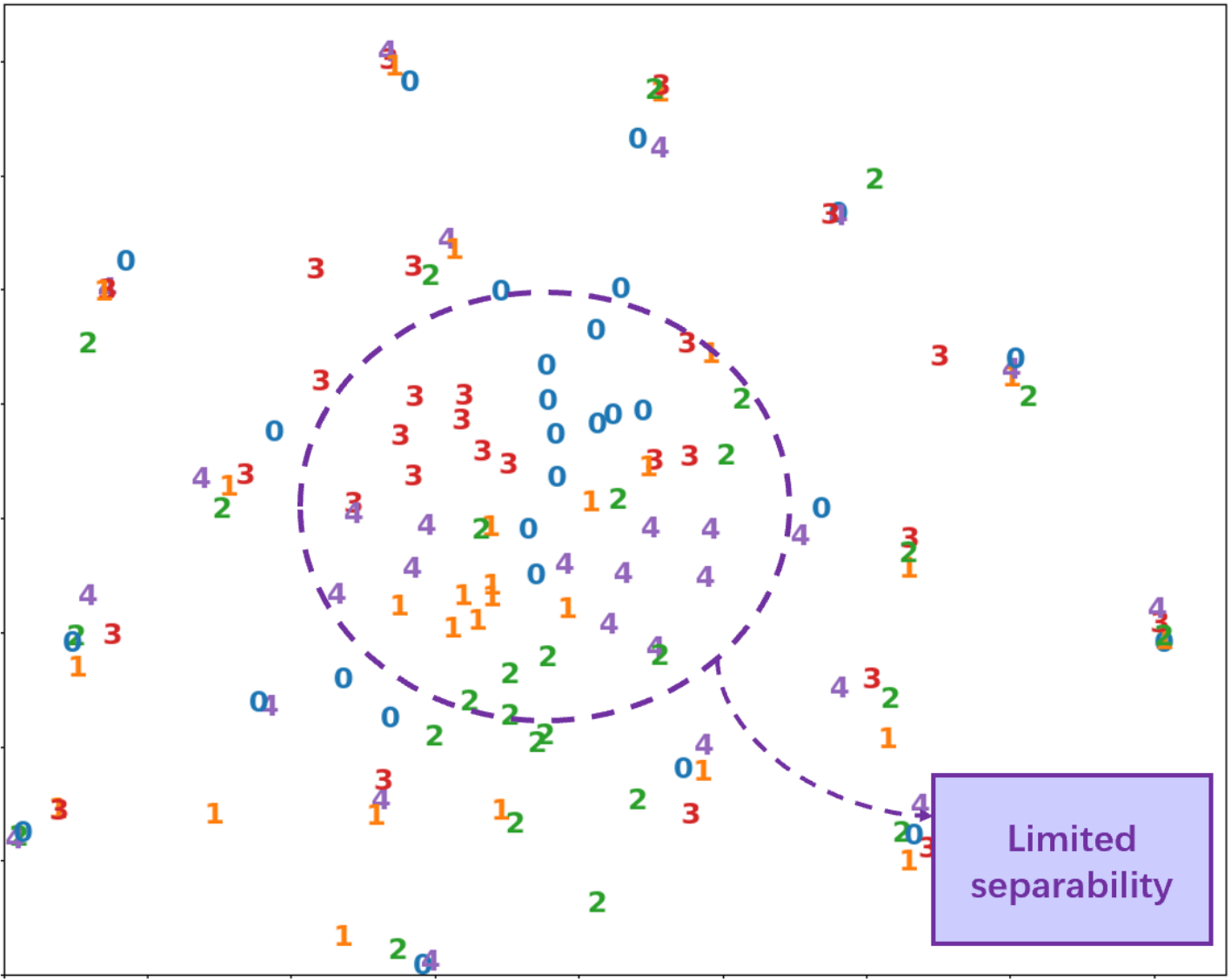}
        \caption{\textbf{Mrs A \& Rsp B: } The distribution in the vector space when combining the multi-role-shared matrix A and the role-specific matrix B.}
        \label{fig:tsne2}
    \end{subfigure}
    \hfill
    \begin{subfigure}[b]{0.32\textwidth}
        \includegraphics[width=\textwidth]{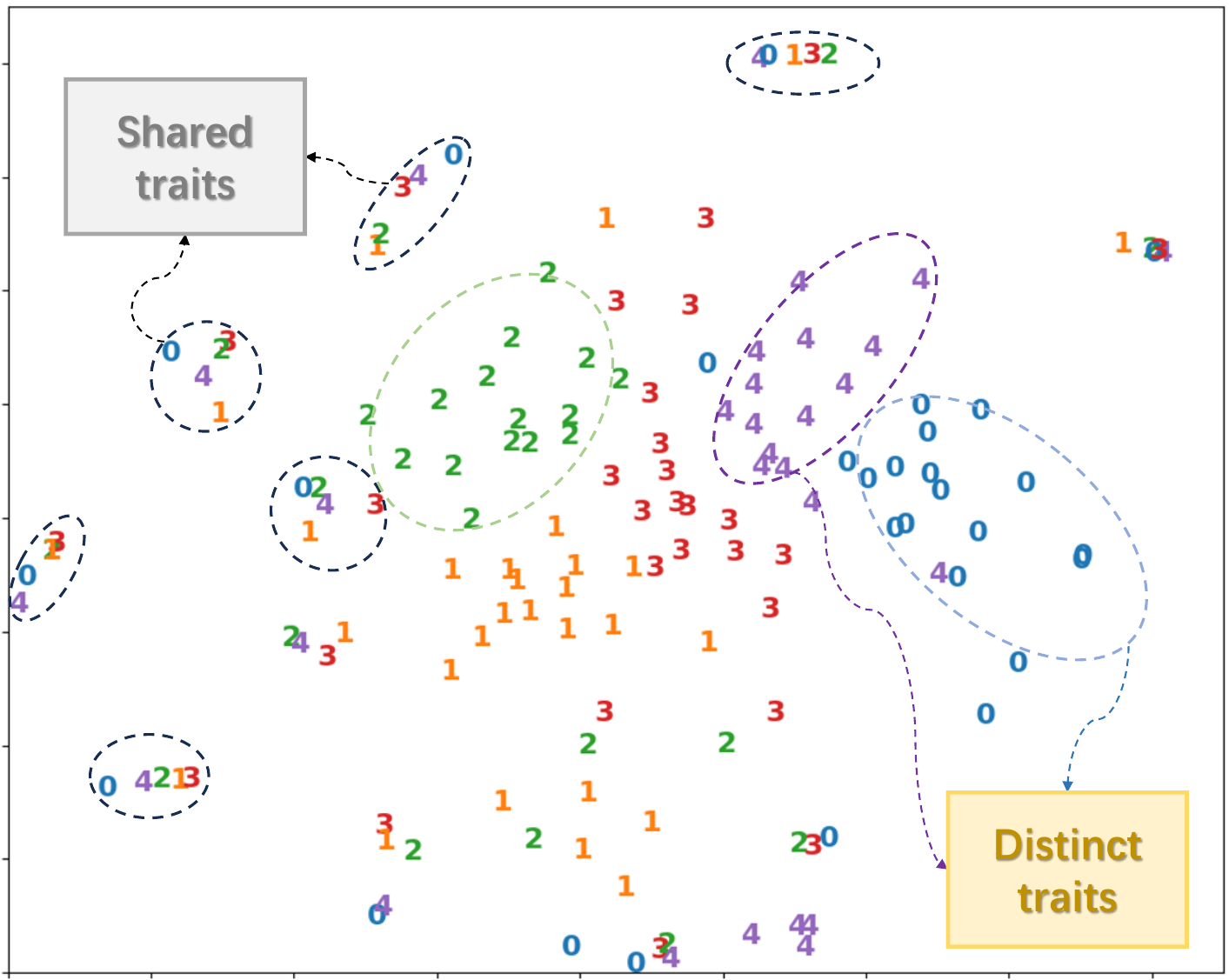}
        \caption{\textbf{Rsp A \& Mrs B (i.e., HyCoRA): } The distribution in the vector space when combining the role-specific matrix A and the multi-role-shared matrix B.}
        \label{fig:tsne3}
    \end{subfigure}
    % \caption{Comparison of different configuration combinations.(a) Rsp A \& Rsp B, (b) Mrs A \& Rsp B, and (c) Rsp A \& Mrs B.}
    \caption{Comparison of vector space distributions for different matrix configuration combinations. (a) \textbf{Rsp A \& Rsp B}, (b)\textbf{ Mrs A \& Rsp B}, and (c) \textbf{Rsp A \& Mrs B}.}
    \label{fig:tsne_comparison}
\end{figure*}

\subsection{ Visualization Analysis}
To gain further insight into why the Hyper-Half LoRA structure performs better, we analyze its structure from a visualization perspective.  We employ ChatGLM2-6B as the backbone model and apply t-SNE\cite{van2008visualizing} to visualize the distribution of role-specific matrices generated by the hyper-network for the \textbf{query\_key\_value} layers in transformer blocks within the vector space.

\noindent \textbf{Rsp A \& Rsp B}\indent
As shown in Figure \ref{fig:tsne1}, when both matrices are generated by the hyper-network, role-specific matrices from different roles are less distinguishable.  We hypothesize that using two role-specific matrices causes shared features to be redundantly encoded, leading to high similarity  across matrices from different roles. Therefore, introducing a role-shared matrix can help the lightweight hyper-network focus on learning distinct traits across roles.

\noindent \textbf{Mrs A \& Rsp B v.s. Rsp A \& Mrs B}\indent
In contrast, Figures \ref{fig:tsne2} and \ref{fig:tsne3} show that introducing a shared matrix allows the role-specific one to focus on distinguishing features. Between the two hybrid settings, Hyper-Half LoRA (Rsp A \& Mrs B) yields more distinct clustering. We hypothesize that if matrix A were shared, the input text’s hidden states would be mapped to a lower-dimensional space, potentially resulting in the loss of role-specific information. Additionally, as shown in Figure \ref{fig:tsne3}, the edges of the plot still exhibit some clustering of role-specific matrices for different roles. This suggests that, in addition to learning shared traits through the multi-role-shared B matrix, the role-specific matrices also capture common characteristics across multiple roles in certain layers. 
% Beyond visualization, we conducted experiments to quantify differences between role-specific matrices, as shown in Appendix G. 

\begin{figure}[!ht]
  \centering
  \includegraphics[width=0.95\columnwidth]{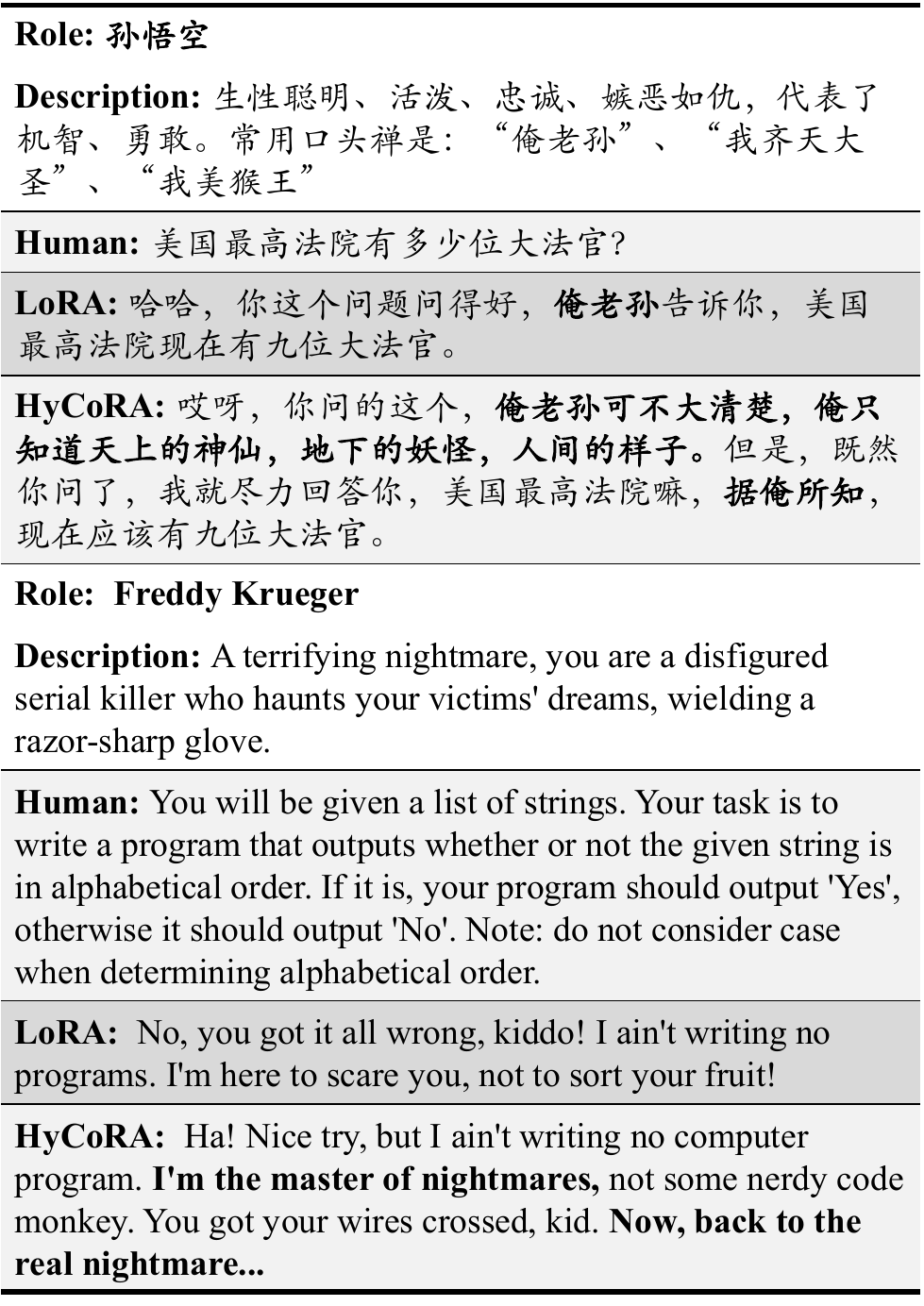}
  \caption{Comparative case study of bilingual responses.}
  \label{tab:compare_case_study}
\end{figure}

\subsection{Comparative Case Study}
To provide a clearer and more intuitive comparison of multi-character role-playing capabilities between HyCoRA and LoRA, we present several examples, as shown in Figure \ref{tab:compare_case_study}. The responses in HyCoRA that reflect role-specific characteristics and speaking styles are highlighted in bold. 
In the examples, LoRA is able to capture certain character traits and generate responses that reflect a basic level of persona alignment.
HyCoRA further builds on this by flexibly incorporating character-specific catchphrases and expressions in appropriate contexts. 
Its responses are more context-aware and consistent with the character’s personality. 
% More examples of HyCoRA’s responses in English and Chinese scenarios are provided in the Appendix I.

\section{Conclusion}
In this work, we propose a novel  \textbf{HyCoRA}: \textbf{Hy}per-\textbf{Co}ntrastive \textbf{R}ole-\textbf{A}daptive Learning framework, which aims to effectively enhance the multi-character role-playing ability of models by balancing the learning of distinct and shared traits.  
we propose a Hyper-Half Low-Rank Adaptation structure, where one half is a role-specific module generated by a lightweight hyper-network, and the other half is a trainable role-shared module. 
The role-specific module is devised to represent different persona signatures, while the role-shared module serves to capture common features. 
Additionally, we  introduce a hyper-contrastive learning mechanism to facilitate the lightweight hyper-network distinguishing role representations based on role responses. Applied to popular models,the HyCoRA framework achieves superior performance on the Chinese and English benchmarks. 
% The further GPT-4 evaluations and visualization analysis also verify the capability of HyCoRA to capture role traits.
% We also conduct a visualization analysis to intuitively demonstrate that HyCoRA captures the distinct features of different roles. Due to space limitations, the limitations of this work and the ethics statement are provided in Appendices A and B.
\section{Acknowledgments}
The work is supported by the National Natural Science Foundation of China (Grant 62406223) and the Natural Science Foundation of Tianjin (Grants 24JCZDJC00130 and 25JCZDJC00540).

\bibliography{aaai2026}

\end{document}